\documentclass{article}
\usepackage[margin=1.4in]{geometry}
\usepackage{graphicx}%
\usepackage{multirow}%
\usepackage{amsmath,amssymb,amsfonts}%
\usepackage{amsthm}%
\usepackage{mathrsfs}%
\usepackage[title]{appendix}%
\usepackage{longtable}
\usepackage{multicol}
\usepackage[table]{xcolor}%
\usepackage{textcomp}%
\usepackage{manyfoot}%
\usepackage{booktabs}%
\usepackage{algorithm}%
\usepackage{algorithmicx}%
\usepackage{algpseudocode}%
\usepackage{listings}%
\usepackage{subcaption}
\usepackage{hyperref}
\usepackage{orcidlink}
\usepackage{authblk}
\usepackage[round,authoryear]{natbib}

\title{\vspace*{-0.1in} Predictive Analytics of Selections of Russet Potatoes}
\author[1,*]{Fabiana Ferracina  \orcidlink{0000-0003-4634-4880}}
\author[1,*]{Bala Krishnamoorthy \orcidlink{0000-0002-2727-6547}}
\author[2]{Mahantesh Halappanavar \orcidlink{0000-0002-2323-4753}}
\author[3]{Shengwei Hu \orcidlink{0009-0003-7150-3617}}
\author[3,*]{Vidyasagar Sathuvalli  \orcidlink{0000-0001-5727-9987}}

\affil[1]{Department of Mathematics and Statistics, Washington State University, 14204 NE Salmon Creek Avenue, Vancouver, 98686, Washington, USA}
\affil[2]{Pacific Northwest National Laboratory,
           902 Battelle Boulevard, 
           Richland,
           99354, 
           Oregon,
           USA}
\affil[3]{College of Agricultural Sciences, Oregon State University,
           2121 South 1st Street, 
           Hermiston,
           97838, 
           Oregon,
           USA}
\affil[*]{Corresponding authors: \href{mailto:fabiana.ferracina@wsu.edu}{fabiana.ferracina@wsu.edu}, \href{mailto:kbala@wsu.edu}{kbala@wsu.edu}, \href{mailto:vidyasagar@oregonstate.edu}{vidyasagar@oregonstate.edu}}

\begin{document}
\maketitle

\begin{abstract}
  We explore the application of machine learning algorithms specifically to enhance the selection process of Russet potato (\emph{Solanum tuberosum} L.) clones in breeding trials by predicting their suitability for advancement. This study addresses the challenge of efficiently identifying high-yield, disease-resistant, and climate-resilient potato varieties that meet processing industry standards. Leveraging manually collected data from trials in the state of Oregon, we investigate the potential of a wide variety of state-of-the-art binary classification models.
  The dataset includes 1086 clones, with data on 38 attributes recorded for each clone, focusing on yield, size, appearance, and frying characteristics, with several control varieties planted consistently across four Oregon regions from 2013-2021.
  We conduct a comprehensive analysis of the dataset that includes preprocessing, feature engineering, and imputation to address missing values.
  We focus on several key metrics such as accuracy, F1-score, and Matthews correlation coefficient (MCC) for model evaluation. The top-performing models, namely a feedforward neural network classifier (Neural Net), histogram-based gradient boosting classifier (HGBC), and a support vector machine classifier (SVM), demonstrate consistent and significant results.
  To further validate our findings, we conducted a simulation study using the aims, data-generating mechanisms, estimands, methods, and performance measures (ADEMP) framework, simulating different data-generating scenarios to assess model robustness and performance through true positive, true negative, false positive, and false negative distributions, area under the receiver operating characteristic curve (AUC-ROC) and MCC.
  The simulation results highlight that non-linear models like SVM and HGBC consistently show higher AUC-ROC and MCC than logistic regression (LR), thus outperforming the traditional linear model across various distributions, and emphasizing the importance of model selection and tuning in agricultural trials. Variable selection further enhances model performance and identifies influential features in predicting trial outcomes.
  The findings emphasize the potential of machine learning in streamlining the selection process for potato varieties, offering benefits such as increased efficiency, substantial cost savings, and judicious resource utilization. Our study contributes insights into precision agriculture and showcases the relevance of advanced technologies for informed decision-making in breeding programs.
\end{abstract}

\section{Introduction}\label{sec1-intro}

Potato (\emph{Solanum tuberosum} L.) is an important food crop in the United States with an annual potato production of $\approx$ 43.9 billion pounds in 2022.
The Pacific Northwest forms the largest sub-region in potato production in the U.S., contributing 60.1\% of U.S.~potato production according to USDA's National Agricultural Statistics Service.
Potatoes are not only consumed as a fresh vegetable but also form a raw material for the French fry, chip, and starch processing industries.
In the U.S., 39\% of the potato crop is utilized by the frozen processing sector.
The Columbia basin of Oregon and Washington states is the leader in the frozen potato sector, with more than 80\% of potatoes grown here being russet type for utilization by the processing industry.

Developing new russet potato varieties with high yield, good processing quality, climate resilience, and disease resistance is the main focus of the Oregon State University’s Potato Breeding and Variety Development program, an integral part of the Northwest Variety Development Program (also known as the Tri-State program).
Typically, it takes 12--13 years from the hybridizations to select and release a new potato variety.
In the first field year, seedling tubers generated from true potato seeds are planted in Klamath Falls (around 60,000 clones) as single hills and about 1\% of the clones selected are planted at Hermiston and Klamath Falls in the second field year.
Phenotypic field selections are performed in the first and second years and then the selected clones are tested through Oregon statewide replicated trials at Hermiston, Klamath Falls, and Ontario.
The Oregon statewide trials are typically performed in years three to five and the selected clones will then move to a Tri-State trial, and then to regional trials from the sixth or seventh field year to the tenth or eleventh year.
Potato varieties released from the Tri-State variety development program such as the ``Clearwater Russet'', ``Mountain Gem Russet'', ``Umatilla Russet'', and ``Ranger Russet'' have proved to be significantly successful by producing higher total yield and higher percentage of finished product derived from raw tubers in the processing industry. Benefits from the adoption of new russet potato varieties continue to grow over time \citep{ArLo2002}.

In the breeding program, the selected potato clones must be grown and evaluated in research replicated trials for multiple years.
Each selected clone will be planted in diverse locations and agronomic data is collected and evaluated for its performance.
Data for each potato clone collected annually includes field data such as emergence, plant vigor, plant uniformity, stems per plant, vine maturity; after harvest data includes total yield, yield for different weight categories; attributes pertaining to the potato's external characteristics, specific gravity, internal defects, and performance following the process of frying.
The extensive costs associated with managing these selection trials and the subsequent data collection after harvest account for a significant proportion of investment in the Oregon Potato Variety Development Program.

The complexity further increases due to the intricate interplay between distinct traits and environmental interactions needed for multi year and multi location trials.
Breeders must decide whether the potato clone should go into the subsequent selection trial stage based on its performance or whether retaining a poorly performing clone could add unnecessary cost to the program resulting in decreased breeding selection efficiency.
Advances in genetic tools have resulted in the use of molecular markers and genomic selection to improve selection efficiency.
Even with molecular and genomic data, it is essential to carry multi year, multi location trials and the breeder must select which clones to advance.
In this context, machine learning technologies can help evaluate potato clones and assist in making such decisions \citep{Dwetal2021}.
The ascendancy of machine learning technologies in revolutionizing data analytics is evident, particularly within breeding programs.
Researchers have used machine learning models to obtain highly accurate predictions of root yield based on the vegetative indices of the different growth stages of cassava ({\it Manihot esculenta Crantz}) \citep{Seetal2020}.
With the potent integration of machine learning algorithms and chemical composition as predictors, researchers have successfully predicted consumer flavor perceptions of blueberries and tomatoes \citep{Coetal2022}.

Data-informed decision making facilitated by machine learning models can assist breeders select potato clones more accurately and reduce the number of varieties proceeding to subsequent stages.
The consequential impact includes substantial financial savings, increased breeding selection efficiency, and judicious resource utilization.


\subsection{Related Work}
\label{sec:Related}

Machine learning has shown promise in revolutionizing the selection process in plant breeding programs.
Several studies have demonstrated the potential of integrating advanced data analytics with traditional breeding practices to enhance efficiency and inform decision-making.
Machine learning methods have been applied to many crops whose information was obtained through various data collection techniques \citep{camargo2009image,akbarzadeh2018plant}.

In a study on cassava (\textit{Manihot esculenta} Crantz), researchers employed machine learning algorithms to predict root yield based on vegetative indices at different growth stages. The findings highlighted the potential of machine learning in providing insights into the association between above and below-ground traits, offering valuable information for breeding programs \citep{Seetal2020}.

Similarly, machine learning methods have been applied to predict consumer flavor perceptions of blueberries and tomatoes. By leveraging chemical composition data and machine learning algorithms, researchers successfully predicted flavor profiles, demonstrating the potential of these techniques in guiding breeding efforts for improved sensory qualities \citep{Coetal2022}.

In a study of French fry potato in Poland, machine learning methods of regression to predict yield using diverse datasets (agronomic, weather, satellite, and soil data) were used  \citep{kurek2023prediction}. The authors found that consideration of features, coupled with appropriate data preprocessing techniques (e.g., imputation) and outlier management, is crucial for accurate predictions in the field of agricultural yield forecasting.

Another study of potato employed the machine learning models of support vector machines (SVM) and multi-layer perceptrons (MLP, also known as Neural Nets) to classify potato varieties based on macroscopic observations and measurements for selection or de-selection during trials much like those used in our study \citep{ozguven229use}. For each of these two classification techniques, four models were constructed---two for binary classification and two for 4-class classification.
The results were reported in terms of the accuracy score (number of correct predictions) with the SVM method demonstrating higher accuracy than MLP in the 4-class models. It is unclear how many features \cite{ozguven229use} considered or how they addressed missing data. Further, no parameter tuning was performed.

These studies collectively highlight the growing application of machine learning in plant breeding programs. By leveraging advanced data analytics, breeders can gain valuable insights, streamline selection processes, and make informed decisions to accelerate the development of improved crop varieties.

\subsection{Contributions of This Study}

This study presents a comprehensive evaluation of various classification models for predicting the selection of potato clones in agricultural trials. Our contributions are multifaceted, encompassing both methodological advancements and practical insights:

\begin{enumerate}
    \item \textbf{Comprehensive Model Comparison:} We systematically compare a diverse set of classification models, including traditional approaches such as Na{\"i}ve Bayes and decision trees, as well as more advanced techniques like multi-layer perceptron classifiers (referred to as Neural Nets), histogram-based gradient boosting classifiers (HGBC), and support vector machines (SVM). This thorough comparison highlights the strengths and weaknesses of each model within the context of an imbalanced agricultural dataset. By understanding these strengths and weaknesses, we aim to improve the accuracy of selecting potato clones with desirable traits such as yield, disease resistance, and processing quality, which are important for successful breeding outcomes.
    \item \textbf{Evaluation Metrics:} By employing the Matthews correlation coefficient (MCC) alongside accuracy and F1-score, we provide a nuanced assessment of model performance. The MCC, in particular, proves to be a valuable metric for evaluating classifiers on imbalanced datasets, offering insights that traditional metrics might overlook. This is particularly important for distinguishing between clones that should be advanced or discarded based on their potential to meet industry standards for yield and quality.
    \item \textbf{Impact of Data Imputation:} We investigate the effects of multiple imputation on model performance, providing a detailed analysis of how imputation influences the predictive power of different classifiers. This analysis underscores the importance of handling missing data appropriately in agricultural datasets. Proper imputation allows us to consider all clones participating in the trials, thus allowing classification models to learn from a more diverse dataset and be more generalizable.
    \item \textbf{Forward Feature Selection:} Through forward feature selection, we identify key variables that significantly impact model performance. This process not only enhances the predictive accuracy of our top-performing models but also offers valuable insights into the most critical factors influencing potato clone selection.
    \item \textbf{Simulation Study:} To validate the robustness of our findings, we conduct a simulation study that replicates various scenarios of data imbalance. This simulation study demonstrates the generalizability of our results and confirms the efficacy of our methodological approach under different conditions.
    \item \textbf{Practical Implications:} Our findings have practical implications for potato breeding programs. By identifying the most influential variables and the best-performing models, we provide actionable recommendations that can enhance the efficiency and accuracy of clone selection processes in agricultural trials. Ultimately, this contributes to developing potato varieties that are better suited to meet market demands and environmental challenges, leading to more sustainable agricultural practices.
\end{enumerate}

Overall, this study contributes to agricultural data science by rigorously evaluating classification models to tackle the specific problem of optimizing potato breeding trials. We highlight how selecting appropriate evaluation metrics can improve model performance in identifying clones with desirable traits such as yield, size, appearance, and frying quality. Furthermore, we demonstrate how advanced machine learning techniques can streamline clone selection processes, ultimately reducing costs and improving resource allocation in breeding programs.


\section{Methods}\label{sec2-mtds}

Code for the methods described here can be found in our repository: \url{https://github.com/fabstat/burbank}.

\subsection{Study Area and Data}
\begin{figure}[ht!]
    \centering
    \includegraphics[width=.7\textwidth]{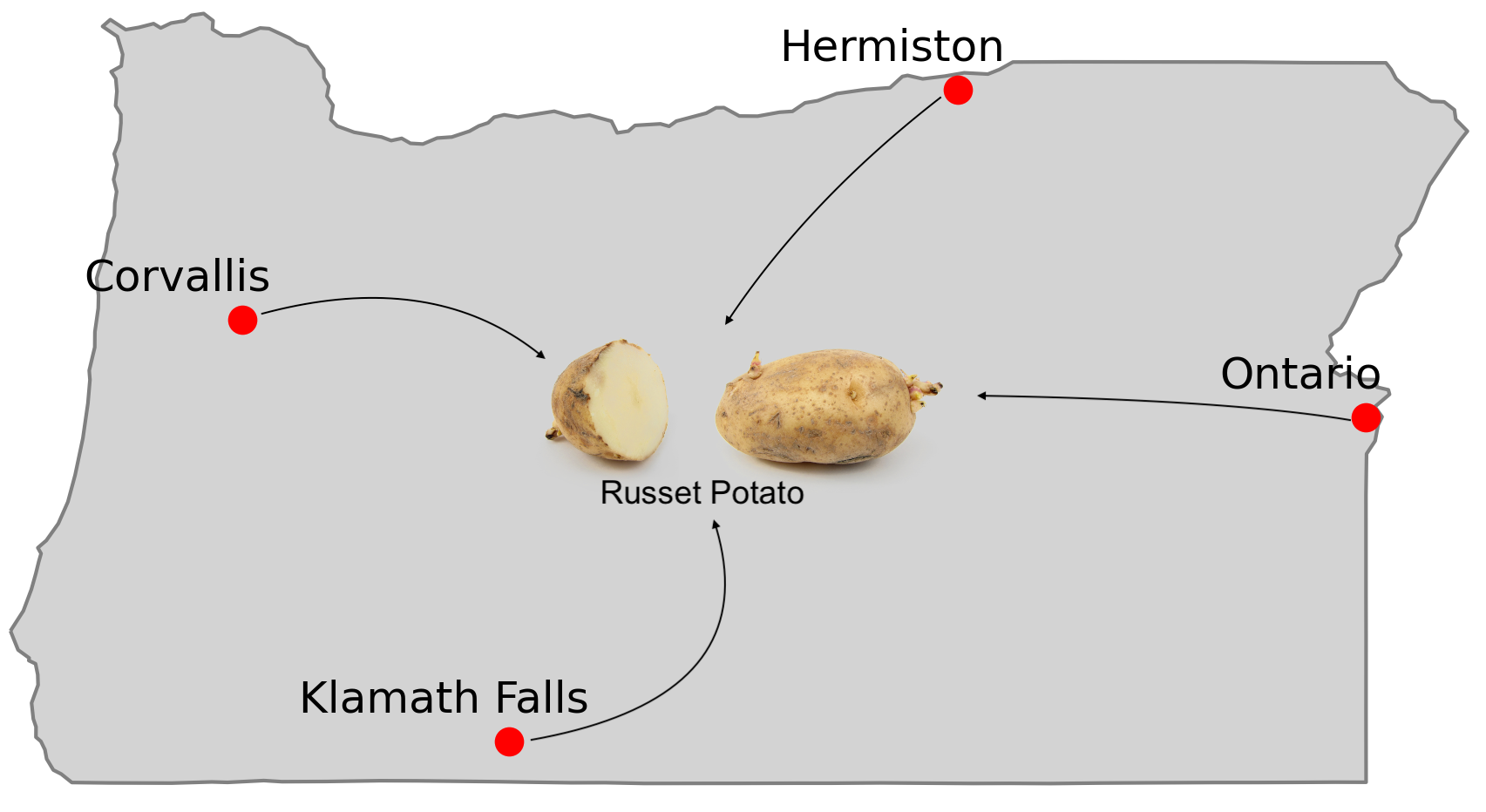}
    \caption{US State of Oregon. Points represent the cities in which the potato trials were conducted: Corvallis, Hermiston, Ontario, and Klamath Falls. From each of these states, russet potatoes such as the one depicted at the center of the map  \citep{potato_pic} were collected and data of each clone's characteristics were recorded for analysis.}
    \label{fig:studyarea}
\end{figure}

From 2013--2014 there were four regions in Oregon, represented in Figure \ref{fig:studyarea}, in which the Russet potato trials were conducted: Hermiston (HER), Ontario (ONT), Klamath Falls (KF) and Corvallis (COR).
Beginning in 2015, COR was dropped from the russet trials and two trials were conducted in HER, early and late (see Table \ref{tab:vine_kill_dates}).

\vspace*{-0.1in}
\begin{table}[ht!]
    \centering
    \caption{Early and late planting season dates over the years. DAP indicates the days after planting for each vine (includes kill date).}
    \label{tab:vine_kill_dates}
    \begin{tabular}{@{}lllllll@{}}
    \toprule
         & \multicolumn{3}{c}{Early Planting} & \multicolumn{3}{c}{Late Planting}  \\\cmidrule{2-4}\cmidrule{5-7}
        
        Year & Planting & Vine Kill & DAP & Planting & Vine Kill & DAP\\
        \midrule
        2013&&&			&04/05/13&	09/04/13 & 152 \\
        
        2014&&&			&04/16/14&	09/11/14 & 149 \\
        
        2015&	03/26/15&	07/22/15&	119 & 04/06/15&	09/04/15 & 152 \\
        
        2016&	03/25/16&	07/29/16& 127 &	04/07/16&	09/02/16 & 149\\
        
        2017&	03/27/17&	08/10/17& 137 & 04/04/17&	08/28/17& 147\\
        
        2018&	03/26/18&	08/08/18& 136 & 04/09/18&	08/31/18& 145\\
        
        2019&	03/28/19&	08/01/19& 127 & 04/16/19&	09/02/19& 140\\
        
        2020&	03/23/20&	08/01/20& 132 & 04/21/20&	08/31/20& 133\\
        
        2021&	03/17/21&	07/30/21& 136 & 03/31/21&	08/31/21& 154\\
        \bottomrule
    \end{tabular}
    
\end{table}

The preprocessed dataset includes 1086 total clones, 216 of which are unique.
Each clone is trialed 1--3 years.
Most clones were in their first year of trial.
There are three control varieties planted in all regions and all years: Ranger Russet, Russet Burbank, and Russet Norkotah.
Two other control varieties are tested only in Hermiston: Shepody (early) and Umatilla (late), and were measured for some but not all years.
For each clone, data for 38 attributes are recorded in a spreadsheet for each particular region.
This data pertained to the total yield, yield per size category, features related to the potato's appearance, and features related to the potato after being fried.
See Table \ref{tab:potato_feat} for more details on each feature.

\subsection{Feature Engineering}

Not all 38  aforementioned features were used in our analysis, and some additional features were included.
Features with over 400 not available (NA) ratings were dropped: ``flower color'', ``vine size'', and ``maturity''.
The feature ``bruise data'' was also dropped as it contained mostly text description and was missing for most observations.
The following new features were added: 
\begin{enumerate}
   \item ``true\_keeps'': response variable. A clone is assigned a ``1'' if it passed three trials and a ``0'' if it passed fewer than three trials.
     Clones in the first or second year of the statewide trials during this analysis were removed from the dataset.
   \item ``Ctrl ave'': Average total yield of control clones for a particular year and trial region.
     All clones from a particular year and trial region should have the same ``control average''.
   \item ``\% CA'': $\displaystyle 100 \times \left(\frac{\textnormal{Total Yield}}{\textnormal{Ctrl Ave}}\right)$ represents the percentage of a particular variety's total yield from the averaged controls' total yield.
    \item ``GDD 1--60'': the sum of the first 60 growing degree days
    \item ``GDD 61--90'': the sum of growing degree days from day 61 to day 90
    \item ``GDD 91--end'': the sum of growing degree days from day 91 to vine kill date
\end{enumerate}
The last three variables pertain to \emph{growing degree days} (GDD) defined as
\[\mathrm{GDD} = \left[\frac{(T_{\mathrm{MAX}} - T_{\mathrm{MIN}})}{2}\right] - T_{\mathrm{BASE}}\]
where $T_{\mathrm{MAX}}$ and $T_{\mathrm{MIN}}$ are the day's maximum and minimum air temperatures, respectively, and $T_{\mathrm{BASE}}$ is ``the temperature below which the process of interest does not progress'' \citep{mcmaster1997growing}.
In our case, $T_{\mathrm{BASE}}$ is the lowest temperature at which potato continues to grow, or 8$^\circ$ Celsius.

\subsection{Data Imputation}\label{ssec:imp}

While the overall dataset was reasonably large, a significant number of variables were missing values (or entries) for several observations.
For instance, the variables ``flower color'' and ``maturity'' were missing for 638 and 639 observations, respectively (of the total of 1086 clones).
When considering only observations with numeric yield, 25 of the 36 original attributes have at least 1 missing missing value, with 13 of these features each having more than 100 missing values.
Hence, imputation was necessary to consider the complete dataset.
Without imputation, one must drop observations with missing values.
Since a few of the attributes were not recorded for almost half of the dataset, dropping observations after data processing and cleaning would reduce the set from 1086 observation to 209 observations. Such a small set of observations would lead to overfitting and machine learning that might not be meaningful.

To impute data, we first standardized the dataset and then used Scikit-learn's \citep{scikit-learn} function \verb+IterativeImputer+ with \verb+sample_posterior=True+, which performs multiple imputation by chained equations, or MICE \citep{van2011mice}.
Under the assumption that the data is missing at random, MICE works by first performing a simple mean imputation to serve as place holders for all variables.
For each variable in turn, the place holder means are removed and the variable is treated as dependent on the other independent variables in a regression problem.

Imputation was done after the train/test/validate data splits, ensuring no information leakage between these sets. We validated our imputation technique by running our prediction models on the subset of the original data containing complete information, and on the same subset with values removed randomly following the distribution of missing values for the overall dataset. There was a slight performance improvement with the imputed set, but overall the results between the two sets remained consistent.

Besides the imputed dataset, we also constructed a non-imputed dataset where we chose a maximum number of ``NA's'' any given column could have so that the dataset was reduced to at most the largest number of NA's that was below this maximum threshold.
In the analysis presented below, an NA threshold of 400 was selected to ensure we kept at least two thirds of the dataset even when dropping rows containing NA's.
For purposes of comparison, we constructed both the imputed and non-imputed datasets by first dropping columns with over 400 NA's.
Postprocessing there were 885 observations in the imputed set and 404 observations in the non-imputed set, with the same set of 40 features in each.

Once numerical variables were standardized, we hot encoded the categorical variables---a process where all categorical variables' values are accounted for all observations, but for an observation's given category, its value is encoded to 1 while the other non-available values are encoded to zero.
Currently our framework does not handle imputation on categorical variables, so observation for which categorical values were missing would be dropped.
However the only categorical variables in our dataset are ``Trial Region'' and ``Year in Trial'', and there are no missing values for them.

We ran a grid search with cross validation to find the best set of hyperparameters for each of the models considered (see Section \ref{ssec:mdlsmtrcs}) on all imputed and non-imputed datasets.

\subsection{Modeling Approach}
With the ultimate goal of automating the process of discriminating varieties that meet or exceed expected targets from those that do not meet the standards specified above,
we explored a large family of machine learning classification models on our dataset and drew structural insights from their results.
Note that this process is currently fully manual, where the potato breeding expert looks at every datum in order to make a decision.
As such, having a more automated way to make, or aid in making, the decision is of great advantage.
We concentrated on nonlinear classifiers since we expected them to perform as well or better than linear ones.
Nonlinear classifiers are more flexible. They do not rely on a line, or hyperplane in higher dimensions, to separate the data classes, and they are more robust against outliers.

Clones must be selected for breeding in all three years of the Oregon statewide trials, if they are to graduate to the Tri-state trials.
We labeled each clone in the dataset as ``1'' if it passed three trials and as ``0'' if it passed fewer trials.
Clones were removed from the dataset if still in the first or second year of statewide trialing during this analysis.
We performed model selection and hyperparameter tuning via grid searches using 5-fold cross-validation and split the data such that 80\% was used in training and 20\% was used for testing. 

\subsection{Classification Models and Evaluation Metrics} \label{ssec:mdlsmtrcs}
Since the goal was to separate two classes in a high dimensional feature space efficiently, we constructed a large family of nonlinear classification models. As these classification methods may not be widely used for agricultural data analysis, we present high level explanations for each of them below.
\begin{enumerate}
   \item {\bfseries $\boldsymbol{k}$-Nearest neighbors:}
     This is a fundamental and widely applied clustering method in unsupervised learning.
     The objective of the algorithm is to partition a dataset into $k$ clusters (for a predefined value of $k$) based on similarity, by iteratively computing the Euclidean distance of each data point and cluster representatives, assigning the data points to clusters, and then moving the cluster representative to be the center of the data points it represents.
     Such an algorithm leads to a local optimum and so it is sensitive to initial conditions and shape assumptions.
     Also, since clustering relies on computing arithmetic means, this algorithm is sensitive to outliers \citep{steinbach2009knn}.
     
   \item {\bfseries Gaussian process based on Laplace approximation:}
     This approach involves placing a Gaussian Process (GP) prior over a latent function, which is then transformed through the logistic function to obtain a prior on the class probability.
     The Laplace approximation is employed for posterior inference, approximating the non-Gaussian joint posterior with a Gaussian distribution.
     With this method, positive training examples will give rise to a positive coefficient for their kernel function, while negative examples will give rise to a negative coefficient, making it analogous to the support vector machine (SVM) model (see Method \ref{item:svm} below).
     However, all features provide information to perform predictions, making it an inefficient algorithm in high dimensional spaces \citep{williams2006gaussian}.
     
   \item {\bfseries Gaussian na{\"i}ve Bayes:}
     This is a na{\"i}ve Bayes method where the likelihood of the features is assumed to be Gaussian.
     A na{\"i}ve Bayes method is a supervised learning technique grounded in the application of Bayes' theorem, which incorporates the ``na{\"i}ve'' assumption of conditional independence among all pairs of features given the class variable $y$ and features $x_1, \dots, x_n$ as follows:
     \begin{align*}
       & P(y | x_1, \dots, x_n) = \frac{P(y)P(x_1, \dots, x_n | y)}{P(x_1, \dots, x_n)}\\
       \Rightarrow & P(y | x_1, \dots, x_n) \propto P(y) \prod\limits_{i=1}^n P(x_i | y).
     \end{align*}
     The predicted class variable maximizes the right-hand side of the proportion, given the relative frequency of the class variable in the training dataset \citep{zhang2004optimality}.
     
   \item {\bfseries Quadratic discriminant analysis:}
     QDA is a supervised learning technique that computes discriminant functions for each class, which are then used to determine the decision boundary between each class.
     Assuming each class density has a multivariate Gaussian distribution and by comparing the log-ratio of two classes posteriors, we can obtain the quadratic discriminant functions
     \[\delta_k(x) = -\frac{1}{2}\log|\boldsymbol\Sigma_k|-\frac{1}{2}(x - \mu_k)^T\boldsymbol\Sigma_k^{-1}(x-\mu_k) + \log \pi_k \,.\] 
     The decision boundary between each pair of classes $k$ and $l$ is the set of $x$ values satisfying the quadratic equation $\delta_k(x) = \delta_l(x)$ \citep{hastie2009elements}.
  
   \item {\bfseries Decision tree:}
     This is a supervised learning approach in which a given vector of explanatory variables $X$ is partitioned into the nodes of a binary tree whose edges represent the mutually exclusive set of outcomes of the parent node.
     Each edge leads to a different input feature until it reaches the leaf nodes, where a class is assigned to the sequence of feature node outcomes; the goal is to group samples with the same label.
     Classification depends on the underlying decision tree structure, which is decided by hyperparameters such as the split criteria and the maximum depth of the tree.
     Although simple to understand, decision trees can be unstable or cause over-fitting depending on the choice of hyperparameters \citep{hastie2009elements}.
     
   \item {\bfseries Adaptive boosting:}
     AdaBoosting is a commonly used ensemble learning approach in which a sequence of weak learners (models that are slightly better than guessing randomly), such as small decision trees, is repeatedly applied to modified versions of the data (i.e., by changing the weights applied to the training observations).
     The predictions from all the weak classifiers are then combined to produce a majority vote prediction \citep{hastie2009elements}.
     
   \item {\bfseries Random forest:}
     This ensemble learning approach is an extension of the bootstrap aggregating (or bagging) method, in which random samples of the training data are selected with replacement and then trained on individually using the base learner. Random forests work similarly by constructing an uncorrelated collection of decision trees.
     Each tree in the ensemble is constructed from training data sampled with replacement, and an exhaustive search through the features is performed in order to obtain the best split in the node set.
     Although this is a robust algorithm and might be selected for this reason, it is more complex and time consuming than other ensemble techniques \citep{breiman2001random}.
     
   \item {\bfseries Multi-layer perceptron (Neural Net):}
     This is a feedforward neural network with at least three layers---an input layer, one or more hidden layers, and an output layer---in which neurons are fully connected with a nonlinear activation function.
     The neural net classifier is trained by backpropagation to optimize a loss function.
     The Broyden – Fletcher – Goldfarb – Shanno (BFGS) method, the optimization algorithm used in this framework, is an iterative method for solving unconstrained nonlinear optimization problems \citep{hinton1990connectionist_2}.
     
   \item {\bfseries Histogram-based gradient boosting (HGB):}
     Gradient boosting algorithms are additive machine learning models that combine many weak learners such as decision trees to form a strong learner.
     As finding the best split points for learning the decision trees via sorting the values of each feature is time and memory intensive, the histogram based algorithm buckets continuous feature values into discrete bins and constructs a histogram, thus reducing the number of unique values to sort \citep{ke2017lightgbm}.
     Several hyperparameters in the histogram-based gradient boosting classifier (HGBC) can be optimized during learning, such as maximum tree depths, maximum number of bins and the $L_2$ regularization parameter, to avoid over-fitting issues while affording good performance.
     
   \item \label{item:svm}
     {\bfseries Support vector machine (SVM) with a radial basis function (RBF) kernel:}
     A linear support vector machine (SVM) identifies the plane that maximizes the distance between itself and the closest set of points (specifically the feature vectors from a set of observations) that belong to the two distinct categories.
    Figure \ref{fig:linearSVM} shows a linearly separable dataset with a linear SVM decision boundary.
    \begin{figure}[ht!]
      \centering
      \begin{subfigure}[b]{0.4\textwidth}
        \includegraphics[width=\textwidth]{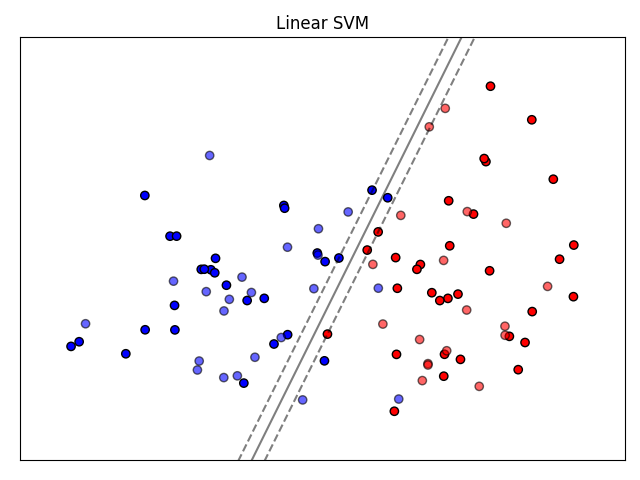}
        \caption{Two classes of points separated by a line.}
        \label{fig:linearSVM}
      \end{subfigure}
      \begin{subfigure}[b]{0.4\textwidth}
        \includegraphics[width=\textwidth]{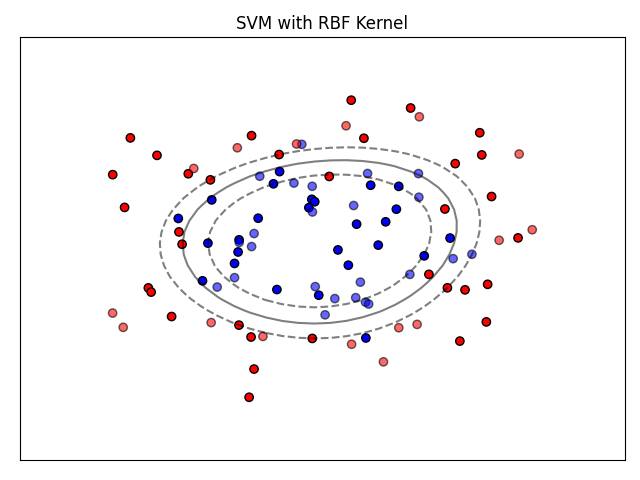}
        \caption{Two classes of points separated by a circular curve.}
        \label{fig:nonlinearSVM}
      \end{subfigure}
      \caption{A dataset where the two classes of points are separable with a line (Left), and one where a nonlinear kernel (radial basis function or RBF in this case) does the separation (Right) when no linear separation is possible.} \label{fig:SVMs}
    \end{figure}
    If the data is not linearly separable for the desired classification problem, then there will be no hyperplane separating the two classes.
    However, a nonlinear decision boundary may still exist.
    As SVM is an inherently linear classification method, we transform the vector space via a \emph{kernel trick}.
    A kernel is a function that take as inputs the observations of the original feature space and outputs the dot products of vectors in a transformed target space where finding hyperplanes that separate the data is more likely \citep{smola1998learning}.
    Two important hyperparameters in a nonlinear SVM are $C$ and $\gamma$.
    The scale parameter $\gamma$ determines the radius of influence of a single observation, while $C$ controls the size of slack variables that allow some observations to be outside the target hyperspace.
    Setting these parameters appropriately is a trade-off between controlling the classification error and overfitting the data.
    
  \item {\bfseries An ensemble of classifiers along with a meta-classifier:}
    Also called a stacking classifier, two or more classifiers are applied to the datasets and the resulting classification predictions are used to train a meta-classifier \citep{tang2014data}.
    The stacking classifier used in this study has gradient boosting and SVM as its level-1 classifiers, and logistic regression (LR) as its level-2 (or meta-) classifier.

\end{enumerate}

\subsubsection{Evaluation Metrics}
To evaluate the quality and performance of the large family of classification models, we computed several metrics including accuracy, F1-score, and Matthews correlation coefficient (MCC).
Each of these metrics provides unique insights into model performance, particularly in the context of imbalanced datasets.

Accuracy is defined as the proportion of correct predictions made by the model across all instances. It is calculated as:
\begin{equation}
\text{Accuracy} = \frac{\text{TP} + \text{TN}}{\text{TP} + \text{TN} + \text{FP} + \text{FN}}
\end{equation}
where TP is the number of true positives, TN is the number of true negatives, FP is the number of false positives, and FN is the number of false negatives. While accuracy is intuitive and widely used, it can be misleading for imbalanced datasets where one class significantly outnumbers the other \citep{powers2011evaluation}.

The F1-score is the harmonic mean of precision and recall, providing a balanced measure of a model's performance. It is particularly useful when dealing with imbalanced datasets \citep{chicco2020advantages}. The F1-score is calculated as:
\begin{equation}
\text{F1-score} = 2 \times \frac{\text{Precision} \times \text{Recall}}{\text{Precision} + \text{Recall}}
\end{equation}
where Precision = TP / (TP + FP) and Recall = TP / (TP + FN). The F1-score ranges from 0 to 1, with higher values indicating better performance.

The Matthews correlation coefficient (MCC) is a measure of the quality of binary classifications, taking into account true and false positives and negatives. It is defined as:
\begin{equation}
\text{MCC} = \frac{\text{TP} \times \text{TN} - \text{FP} \times \text{FN}}{\sqrt{(\text{TP}+\text{FP})(\text{TP}+\text{FN})(\text{TN}+\text{FP})(\text{TN}+\text{FN})}}
\end{equation}
The MCC ranges from $-1$ to $1$, where $1$ represents a perfect prediction, $0$ indicates no better than random prediction, and $-1$ represents total disagreement between prediction and observation.
On imbalanced datasets, MCC can be more informative than either accuracy or the F1-score as it provides a more balanced measure of the proportion of all classes predicted correctly \citep{chicco2017ten, boughorbel2017optimal}.

Our dataset is significantly imbalanced, including 88 positive examples and 998 negative examples. In such cases, the use of MCC is particularly relevant as it takes into account the ratio of the confusion matrix size \citep{chicco2020advantages}. This characteristic makes MCC especially suitable for evaluating binary classification performance on imbalanced datasets, providing a more reliable performance metric compared to accuracy or F1-score alone.

\subsection{Variable Selection}
  We performed variable selection on both the imputed and non-imputed datasets for the top-performing classification models used in this study.
  For each model described in Section \ref{ssec:top}, we performed a grid search with cross-validation on the hyperparameter space to determine the best scoring model.
  We then performed forward feature selection by beginning with zero features in our models and iterating through each available feature, computing each model's score with that feature included in it.
  The feature resulting in the highest model score becomes a feature in the model, and the search for a second feature proceeds.
  The process continues until we have the desired number of features in the model or an additional feature does not significantly improve the model score.
  We validate the feature selection process using 5-fold cross-validation.
  For our study, we explore different fractions of features to select: 0.1, 0.3, 0.5, 0.7, and 0.9, and compare the respective results---we present them in Section \ref{ssec:top}.

\subsection{Simulation Study}

Simulation studies are essential for evaluating the performance of statistical and machine learning models under various conditions. The ADEMP (\underline{A}ims, \underline{D}ata-generating mechanisms, \underline{E}stimands, \underline{M}ethods, \underline{P}erfomance measures) framework, as outlined by \cite{morris2019using}, provides a structured approach to designing and reporting simulation studies. The present study leveraged the ADEMP framework to evaluate the generalizability of the top performing classification models (see \ref{ssec:top}) on simulated datasets that are independent from the original potato trials dataset.

\subsubsection{Aims}
The primary aim of this simulation study was to assess the performance of a subset of classification models used in the overall study, whose goal was to predict whether a potato variety will graduate to the next set of trials (according to binary response variable: \texttt{true\_keeps} used in supervised training). We focused on the top three performing models, and added logistic regression (linear model) for baseline comparison. Focusing on these models helped us keep computation manageable and enabled us to use a comprehensive cross validated grid search approach for each trial.

Specifically, we aimed to:
\begin{enumerate}
    \item Evaluate the predictive accuracy of non-linear models (Neural Network, Histogram-based Gradient Boosting Classifier (HGBC), and Support Vector Machine (SVM)) compared to a linear model (Logistic Regression, LR).
    \item Determine the best-performing model in terms of True Positives (TP), False Positives (FP), False Negatives (FN), True Negatives (TN), Area Under the Receiver Operating Characteristic Curve (AUC-ROC), and Matthew's Correlation Coefficient (MCC).
    \item Measure results variability between each simulated dataset by computing the standard error (SE) of their MCC scores.
    \item Investigate how variations in data-generating mechanisms affect model performance.
    \item Identify and explain both expected and unexpected effects of different data distributions on model performance.
\end{enumerate}

\subsubsection{Data-Generating Mechanisms}
In simulation studies, the choice of data-generating mechanisms is crucial for evaluating the performance of statistical methods under various conditions. For example, \cite{dasgupta2013comparison} selected specific data-generating mechanisms in their study to comprehensively assess the robustness of different clustering methods. They used normal, binomial, Poisson, and Gamma distributions to reflect the diverse characteristics of their protein data and to test how clustering methods handle different distributional properties and data structures.

Similarly, in our study, we varied the data-generating distributions to simulate different scenarios. The original potato variety trials dataset (empirical dataset) contains measurements on potato clones, including weight (or mass) variables, counts, and ratings. These data types were taken into account in our simulation of different scenarios:
\begin{enumerate}
    \item \textbf{Scenario 1:} $\mathrm{Normal}(\mu_{\text{data}}, \sigma_{\text{data}})$ for weight/mass type quantities, $\mathrm{Poisson}(\mu_{\text{data}})$ for count type quantities, $\mathrm{Uniform}(a=\min_{\text{data}}, b=\max_{\text{data}})$ for rating type quantities.
    \item \textbf{Scenario 2:} $\mathrm{Normal}(\mu_{\text{data}}, \sigma_{\text{data}})$ for weight/mass type quantities, $\mathrm{Poisson}(\mu_{\text{data}})$ for count type quantities, $\mathrm{Beta}(a=2, b=5)$ for rating type quantities.
    \item \textbf{Scenario 3:} $\mathrm{Gamma}(k=2, \theta=2)$ for weight/mass type quantities, $\mathrm{Poisson}(\mu_{\text{data}})$ for count type quantities, $\mathrm{Uniform}(a=\min_{\text{data}}, b=\max_{\text{data}})$ for rating type quantities.
    \item \textbf{Scenario 4:} $\mathrm{Gamma}(k=2, \theta=2)$ for weight/mass type quantities, $\mathrm{Poisson}(\mu_{\text{data}})$ for count type quantities, $\mathrm{Beta}(a=2, b=5)$ for rating type quantities.
\end{enumerate}

We applied normal and gamma distributions for weight/mass variables, Poisson distributions for count variables, and both uniform and beta distributions for rating variables across different scenarios. This approach ensured a thorough examination of the classification methods' performance in handling different shapes and types of data, thereby providing a robust evaluation framework. The selection of these mechanisms allowed the capturing of inherent distributional characteristics present in the original dataset, thus enhancing the reliability and generalizability of the study's findings.

For each scenario, 500 datasets were simulated to account for variability. Although a larger number of datasets would be best to fully confirm robustness, applying cross-validated parameter optimization grid search to each classification model incurs significant computational cost. Available computational resources limited our odyssey to 2001 datasets (4 $\times$ 500 simulated + original), each of 885 observations and 41 features. Due to the sample size, and in combination with the other methods employed, we believe this demonstrates a reasonable level of robustness. We leave the task of more intensive computational verification for future study and present our current insights.

\subsubsection{Estimands}
The primary estimands in the simulation study were:
\begin{itemize}
    \item The probability that a potato variety graduates to the next trial set (i.e., \(\mathbb{P}(\texttt{true\_keeps} = 1)\)).
    \item The classification decision for each potato variety (i.e., predicted \texttt{true\_keeps} values).
\end{itemize}

\subsubsection{Methods in Simulation Study}
Four classification models were used to predict the binary response variable \texttt{true\_keeps}:
\begin{enumerate}
    \item Multi-layer Perceptron Classifier (Neural Net)
    \item Histogram-based Gradient Boosting Classifier (HGBC)
    \item Support Vector Machine Classifier with RBF Kernel (SVM)
    \item Logistic Regression (LR)
\end{enumerate}
Each model underwent hyperparameter tuning via cross-validated grid search. Data imputation was performed, where the missingness from the original dataset was mimicked in the simulated datasets prior to imputation. All variables with no more than 400 missing values were considered under each model (i.e. there was no variable selection in the simulation study).

\subsubsection{Performance Measures}
The performance of the models was evaluated using the following metrics:
\begin{itemize}
    \item True Positives (TP)
    \item False Positives (FP)
    \item False Negatives (FN)
    \item True Negatives (TN)
    \item Area Under the Receiver Operating Characteristic Curve (AUC-ROC)
    \item Matthew's Correlation Coefficient (MCC)
    \item Standard error (SE) between MCC scores
\end{itemize}
For each scenario, the metrics were averaged over the 500 simulated datasets.

\section{Results}
\label{sec:Results}

Table \ref{tab:classy} presents the MCCs for the classification models explored in this study on the imputed and non-imputed datasets.
Columns containing more than 400 NA's were dropped before imputation and data cleaning, so that both the imputed and non-imputed datasets contained the same set of features.
The top performing classifiers (highlighted in gray) were the feedforward neural network classifier (Neural Net), histogram-based gradient boosting (HGBC), and support vector machine (SVM), as each obtained an MCC score above or very close to 0.5 in the cross-validation and test for both imputed and non-imputed datasets with high significance.
We note that AdaBoosting achieved similar performance to the declared top performing models, however we will not discuss it further as we will focus on HGBC, which uses the same underlying principle of an ensemble of weak learners but it is a more modern algorithm allowing additional hyperparameters to be set (e.g., regularization).

Models whose test MCC is very close to zero are no better than a random guess.
Gaussian process, Na{\"i}ve Bayes, QDA, and Decision tree classifiers all performed particularly poorly on our datasets.
Although Gaussian process is analogous to SVM, its implementation used in this analysis is non-sparse, and hence the model uses all features for each example when computing the high-dimensional posterior.
As such, a possible explanation for this model's poor performance is that the posterior computed from the training examples is skewed.
If the posterior is skewed, the Laplace approximation (which is centered on the mode) will not be accurate \citep{rasmussen2006gaussian}. 

The poor performance from Na{\"i}ve Bayes and QDA may be due to the assumption of feature independence inherent to these models.
An inspection of the table of features (Table \ref{tab:potato_feat}) makes it clear that several variables considered are not independent.
Lastly, a decision tree's performance is very sensitive to its structure, where features are greedily made into nodes.
It is possible that variable selection and a more thorough search of the hyperparameter space could allow these models to better perform on our dataset.
However, here we focus on those models showing better promise under the same circumstances.

\begin{table}
    \centering
    \caption{Best test Mathews correlation coefficients (MCC) and their respective 95\% confidence intervals (CI) for all the classification models considered in this study.
      Top performing models (highlighted in gray) were the feedforward neural network classifier (Neural Net), histogram-based gradient boosting (HGBC) and support vector machine (SVM), on datasets with both multiple imputation and without imputation.
      The results presented here are from models using all available features which have less than 400 NA's.
      QDA stands for Quadratic Discriminant Analysis.
    }
    \label{tab:classy}
    \begin{tabular}{@{}lcccc@{}}
    \toprule
      & \multicolumn{2}{c}{Imputed} & \multicolumn{2}{c}{Not Imputed} \\
    \cmidrule{2-3} \cmidrule{4-5}
        Model &   Test MCC & CI Test MCC &  Test MCC & CI Test MCC\\
        \midrule
        $k$-nearest neighbors &0.413&(0.378, 0.447)&0.338&(0.283, 0.392)\\
        
        Gaussian process & 0.000 &($-$0.042, 0.042)&0.000&($-$0.062, 0.062)\\
        
        Na{\"i}ve Bayes &0.373&(0.336, 0.408)&$-$0.004&($-$0.066, 0.058)\\
        
        QDA &0.288& (0.250, 0.326)&0.000&($-$0.062, 0.062)\\
        
        Decision tree &$-$0.035&($-$0.077, 0.006)&0.089&(0.027, 0.150)\\
        
        AdaBoosting &0.523&(0.492, 0.553)&0.541&(0.496, 0.583)\\
        
        Random forest &0.085&(0.043, 0.126)&0.531&(0.486, 0.574)\\
        
        \rowcolor{lightgray} Neural Net &0.530&(0.500, 0.559)&0.623&(0.584, 0.659)\\
        
        \rowcolor{lightgray} HGBC  &0.563&(0.534, 0.591)&0.485&(0.436, 0.531)\\
        
        \rowcolor{lightgray} SVM  &0.563&(0.534, 0.591)&0.692&(0.658, 0.723)\\
       
        Stack  &0.534&(0.504, 0.563)&0.432&(0.380, 0.481)\\
        \bottomrule
    \end{tabular}
\end{table}

  Based on the results presented in Table \ref{tab:classy}, if we had to pick three models, we should choose the Neural Net, Histogram-based Gradient Boosting (HGBC) and Support Vector Machine (SVM) models. These three models are highlighted in gray in the table as the top-performing models. Here's the rationale for selecting these models:\\
    \emph{Neural Net:}
    \begin{itemize}
        \item[-] Demonstrates strong performance on both datasets
        \item[-] Has the second-highest MCC score (0.623) for the non-imputed dataset
        \item[-] Shows improvement when moving from imputed (0.530) to non-imputed (0.623) data
    \end{itemize}
    \emph{HGBC:}
    \begin{itemize}
        \item[-] Performs well on the imputed dataset with an MCC of 0.563
        \item[-] While its performance on the non-imputed dataset is lower (0.485), it's still among the top performers overall
    \end{itemize}
     \emph{SVM:}
    \begin{itemize}
        \item[-] Performs exceptionally well on both imputed and non-imputed datasets
        \item[-] Has the highest MCC score (0.692) for the non-imputed dataset
        \item[-] Shows consistent performance across both datasets (0.563 for imputed, 0.692 for non-imputed)
    \end{itemize}
These three models consistently outperform the others across both imputed and non-imputed datasets. They also offer a good variety of approaches (kernel-based, neural network, and ensemble method), which could be valuable for comparing and combining results in an ensemble approach.\\
It's worth noting that AdaBoosting also performs well, especially on the non-imputed dataset (0.541). However, HGBC, which is a more advanced boosting technique, generally outperforms it, especially on the imputed dataset.

\subsection{Top Performing Models} \label{ssec:top}
\begin{figure}[ht!]
    \centering
    \begin{multicols}{2}
    \includegraphics[width=.9\linewidth]{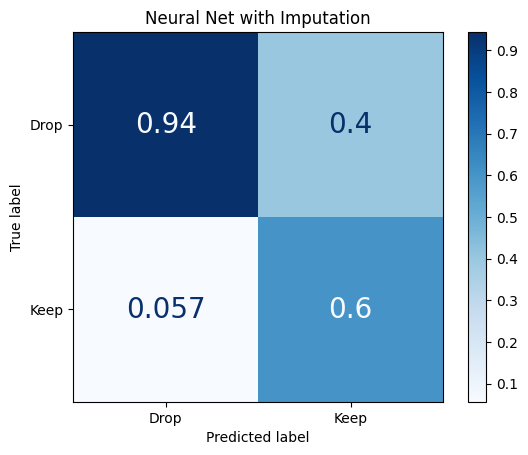}\par
    \includegraphics[width=.9\linewidth]{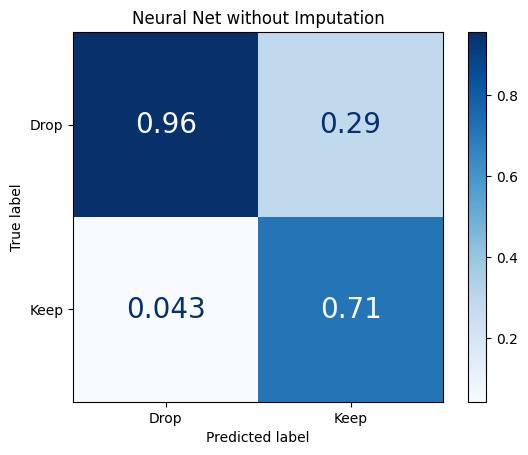}\par
    \end{multicols}
    \vspace*{-0.25in}
    \begin{multicols}{2}
    \includegraphics[width=.9\linewidth]{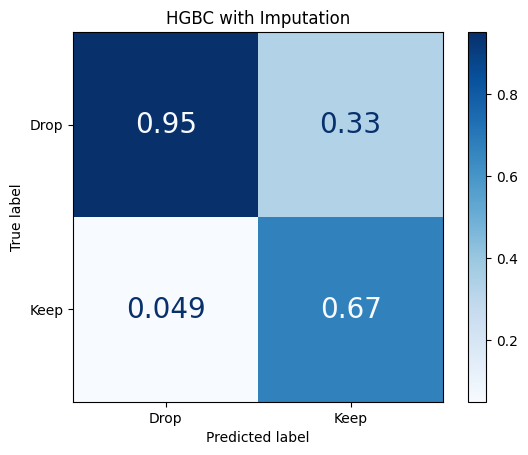}\par
    \includegraphics[width=.9\linewidth]{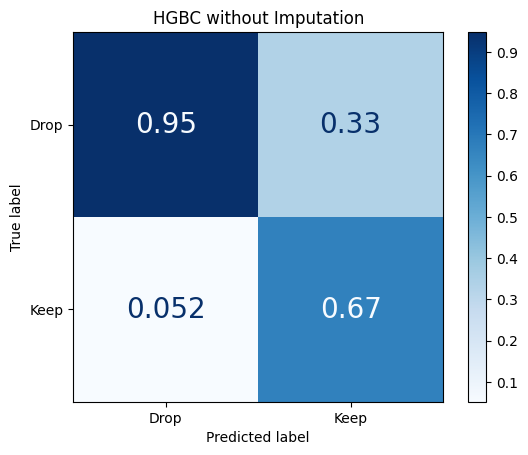}\par
    \end{multicols}
    \vspace*{-0.25in}
    \begin{multicols}{2}
    \includegraphics[width=.9\linewidth]{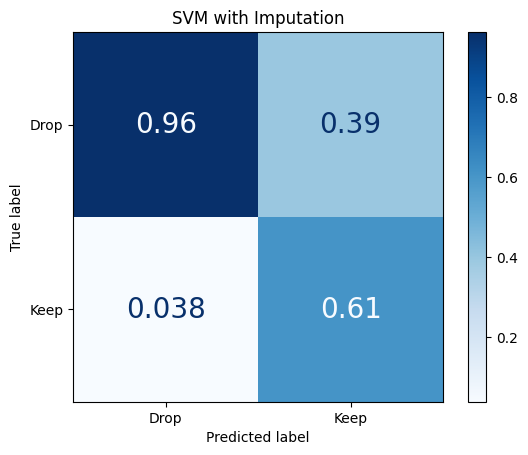}\par
    \includegraphics[width=.9\linewidth]{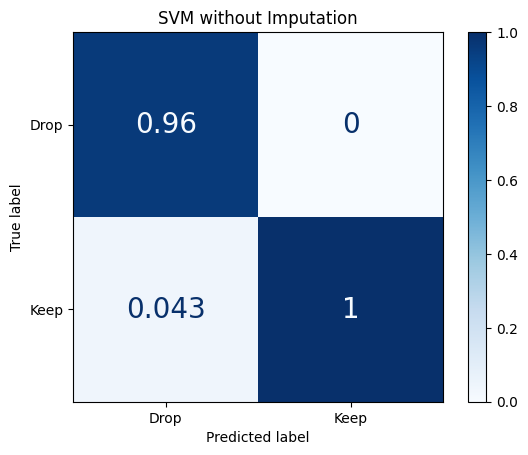}\par
    \end{multicols}
    \vspace*{-0.1in}
    \caption{Confusion matrices for the top performing models applied to both imputed and non-imputed datasets: feedforward neural network classifier (Neural Net) in first row, histogram-based gradient boosting classifier (HGBC) in middle row, and support vector machine (SVM)  classifier in the last row.
      Variable selection was not performed here, however variables containing at least 400 NA's  were dropped before training.
      Numerical values within the confusion matrices represent prediction proportions.}
    \label{fig:cm}
\end{figure}

From among the large family of classification models considered (Section \ref{ssec:mdlsmtrcs}), three models performed consistently well according to their MCC scores, with both imputed and and non-imputed data.
They were the feedforward neural network classifier (Neural Net), the histogram-based gradient boosting classifier (HGBC), and the support vector machine classifier (SVM).
The results from these models without variable selection are highlighted in Table \ref{tab:classy} in light gray, and their confusion matrices are presented in Figure \ref{fig:cm}.

The Neural Net model shows strong performance on both imputed and non-imputed datasets:
\begin{itemize}
    \item[-] High true negative rate (0.94) and true positive rate (0.60).
    \item[-] Similar performance, with a slightly higher true positive rate (0.71) but lower true negative rate (0.96).
\end{itemize}
This suggests that the Neural Net is effective at both identifying varieties to keep and those to drop, with a slight improvement in identifying varieties to keep when using non-imputed data. The HGBC model demonstrates consistent performance across both datasets:
\begin{itemize}
    \item[-] High true negative rate (0.95) but lower true positive rate (0.67).
    \item[-] Nearly identical performance to imputed data.
\end{itemize}
HGBC shows strong ability to identify varieties to drop but is less effective at identifying varieties to keep. The consistency between imputed and non-imputed datasets suggests that HGBC is robust to the imputation process. The SVM model shows interesting differences between imputed and non-imputed datasets:
\begin{itemize}
    \item[-] High true negative rate (0.96) and moderate true positive rate (0.61).
    \item[-] Perfect true positive rate (1.00) but slightly lower true negative rate (0.96).
\end{itemize}
SVM demonstrates excellent performance, particularly on the non-imputed dataset where it perfectly identifies varieties to keep. However, this comes at a slight cost to its ability to identify varieties to drop.

Comparing the three models we can see that all models show strong performance in identifying varieties to drop (high true negative rates). SVM on non-imputed data shows the best performance for identifying varieties to keep, but may be overfitting. Neural Net and HGBC show more balanced performance across imputed and non-imputed datasets. Imputation appears to have a minimal impact on HGBC, moderate impact on Neural Net, and significant impact on SVM performance.

These results suggest that while all three models are effective for potato variety selection, the choice of model and whether to use imputation may depend on the specific goals of the breeding program (e.g., prioritizing identification of varieties to keep vs. varieties to drop).

We also explored forward feature selection on these three models to understand which variables are important in determining whether a potato clone is kept in the trials.
The resulting accuracy, F1 and MCC scores for the top three performing models using forward feature selection with 10, 30, 50, 70 and 90 percent of features included on the imputed and non-imputed datasets are presented in Table \ref{tab:classy-var}.

\begin{table}[ht!]
  \centering
  \caption{
    Performance metrics: Accuracy, F1-score, and Matthews Correlation Coefficient (MCC) for the top classification models after forward feature selection. All reported scores were statistically significant and within 95\% confidence intervals.}
    \label{tab:classy-var}
    \begin{tabular}{@{}llcccccc@{}}
      \toprule
      \% of & Model & \multicolumn{3}{c}{Imputed} & \multicolumn{3}{c}{Not Imputed} \\
      \cmidrule(lr){3-6} \cmidrule(l){6-8}
      features & & Accuracy & F1-score & MCC & Accuracy & F1-score & MCC \\
      \midrule
      \multirow{3}{*}{\shortstack[l]{0.1}} 
      & Neural Net & 0.920 & 0.680 & 0.429 & 0.950 & 0.770 & 0.616 \\
      & HGBC & 0.850 & 0.640 & 0.300 & 0.930 & 0.780 & 0.564 \\
      & SVM & 0.930 & 0.740 & 0.513 & 0.920 & 0.480 & 0.000 \\
      \midrule
      \multirow{3}{*}{\shortstack[l]{0.3}}
      & Neural Net & 0.890 & 0.710 & 0.423 & 0.930 & 0.720 & 0.440 \\
      & HGBC & 0.870 & 0.690 & 0.400 & 0.930 & 0.800 & 0.604 \\
      & SVM & 0.930 & 0.740 & 0.513 & 0.920 & 0.480 & 0.000 \\
      \midrule
      \multirow{3}{*}{\shortstack[l]{0.5}}
      & Neural Net & 0.900 & 0.740 & 0.491 & 0.940 & 0.820 & 0.636 \\
      & HGBC & 0.890 & 0.720 & 0.443 & 0.960 & 0.860 & 0.716 \\
      & SVM & 0.900 & 0.730 & 0.472 & 0.920 & 0.480 & 0.000 \\
      \midrule
      \multirow{3}{*}{\shortstack[l]{0.7}}
      & Neural Net & 0.900 & 0.730 & 0.472 & 0.960 & 0.840 & 0.697 \\
      & HGBC & 0.890 & 0.720 & 0.433 & 0.910 & 0.690 & 0.373 \\
      & SVM & 0.900 & 0.740 & 0.491 & 0.930 & 0.650 & 0.432 \\
      \midrule
      \multirow{3}{*}{\shortstack[l]{0.9}}
      & Neural Net & 0.920 & 0.780 & 0.559 & 0.940 & 0.780 & 0.569 \\
      & HGBC & 0.910 & 0.740 & 0.488 & 0.930 & 0.720 & 0.440 \\
      & SVM & 0.900 & 0.760 & 0.536 & 0.940 & 0.720 & 0.531 \\
      \bottomrule
    \end{tabular}
\end{table}

Overall, model performance generally improves as more features are included, with the best results often seen at 90\% of features, or with no feature selection. The impact of imputation varies across models and metrics, suggesting that the effectiveness of imputation is model and data-dependent. Notably, no single model consistently outperforms the others across all scenarios, indicating that model selection should be context-specific.

The Neural Network shows consistent improvement with more features in both imputed and non-imputed datasets. It achieves the highest accuracy (0.96) and F1-score (0.84) on the non-imputed dataset at 70\% feature selection, demonstrating robust performance across different feature selection stages. 

The HGBC model exhibits variable performance across feature selection stages. It peaks in performance on the non-imputed dataset at 50\% feature selection, with the highest accuracy (0.96) and F1-score (0.86) among all models, but shows more sensitivity to feature selection compared to other models. 

SVM displays consistent performance in the imputed dataset across feature selection stages but notably underperforms in the non-imputed dataset for lower feature percentages, with MCC of 0 at 10\%, 30\%, and 50\% feature selections. However, it improves significantly in the non-imputed dataset with higher feature percentages.

The impact of imputation varies among the models. The Neural Network generally benefits from imputation at lower feature percentages but shows comparable or better performance without imputation at higher percentages. HGBC shows mixed results, with imputation being beneficial in some cases and detrimental in others. SVM significantly benefits from imputation, especially at lower feature percentages where it struggles with non-imputed data.

As forward feature selection becomes more effective with almost no selection, it does not seem beneficial for these data wherein most features appear to contribute valuable information. That said, good performance can also be achieved at lower percentages of feature inclusion (e.g., 50\% and 70\%), indicating some potential for effective dimensionality reduction.

Not surprisingly, yield components emerged as the most important factors, with variables like ``Total Yield'', ``\% RB'', ``Yield No. 1s'', ``\% 1s'', and ``Over 20 oz.'' consistently selected across all models and datasets. Quality traits also played a substantial part in the classification process with features such as ``Specific Gravity'', ``Fry Color'', ``Hollow Heart'', and ``Brown Center'' frequently chosen by the models. Environmental factors, particularly Growing Degree Days (GDD) variables, were also often included in the selected features.

Trial information proved to be another critical aspect, with ``Year in trial'' and ``Trial Region'' consistently appearing in the selected variables. This indicates that the stage of the trial and the location where it's conducted significantly influence the classification results. Interestingly, variables related to defects, such as ``Blackspot'', ``Internal Brown Spot'', and ``Vascular'', were more frequently selected in HGBC models, suggesting this particular algorithm may be more attuned to these quality issues.

The forward variable selection process showed notable consistency in the variables selected across imputed and non-imputed datasets. Many of the same variables were chosen regardless of the dataset type, indicating robustness in variable importance. However, there were some model-specific preferences, with different models prioritizing slightly different sets of variables. This variation could be attributed to the unique algorithms of each model and how they handle feature interactions.

As the analysis progresses to smaller subsets of variables, a clear pattern emerges. The models increasingly focused on core yield components, key quality traits, and essential trial information. This dimensionality reduction suggests that these features are the most critical for accurate potato classification. Overall, this comprehensive examination of variable importance provides valuable insights into the factors that most significantly influence potato classification, offering potential applications in breeding programs and crop management strategies.

As discussed in Section \ref{ssec:imp}, we validated our imputation technique by comparing the results of running our predictive models on a complete dataset and on the same dataset with values randomly removed. As with our main results, columns containing 400 or more NA's were removed, leaving a total of 885 observations, 404 of which were complete. Considering the entire dataset, we computed the distributions of NA's for each feature. We then randomly removed values from each feature in the 404-observations set according to their NA distributions. Table \ref{tab:imp-val}, provides the performance measures obtained from the top three models on each of these sets.

\begin{table}[ht!]
    \centering
     \caption{Performance measures of the top three models on the complete dataset of 404 observations and the same dataset with randomly removed values following the NA distributions of the entire dataset of 885 total observations.}
    \label{tab:imp-val}
    \begin{tabular}{@{}lcccccc@{}}
    \toprule
      & \multicolumn{3}{c}{Imputed} & \multicolumn{3}{c}{Not Imputed} \\
    \cmidrule{2-4} \cmidrule{5-7}
        Model &  Accuracy & F1-score & MCC & Accuracy & F1-score & MCC\\
        \midrule
        Neural Net & 0.97& 0.88 & 0.766& 0.95& 0.80 & 0.623\\
        
        HGBC & 0.94& 0.75 & 0.541& 0.93 & 0.73 & 0.485\\
        
        SVM  & 0.97& 0.88 & 0.766& 0.96& 0.82 & 0.692\\
        \bottomrule
\end{tabular}
\end{table}

\subsection{Simulation Study Results} \label{ssec:simres}

\begin{table}[ht!]
\centering
\caption{Performance metrics for original and simulated datasets. Note that ROC refers to the AUC-ROC score.}
\label{tab:results}
\begin{tabular}{lcccccccc}
\toprule
\textbf{TRIAL} & \textbf{MODEL} & \textbf{TP} & \textbf{FP} & \textbf{FN} & \textbf{TN} & \textbf{ROC} & \textbf{MCC} & \textbf{SE}\\
\midrule
original & Neural Net & 136 & 10 & 7 & 8 & 0.73242 & 0.428767 & -- \\
original & HGBC & 143 & 3 & 10 & 5 & 0.656393 & 0.418387 & -- \\
original & SVM & 142 & 4 & 8 & 7 & 0.719635 & 0.506071 & -- \\
original & LR & 132 & 14 & 6 & 9 & 0.752055 & 0.418739 & -- \\
normal\_uniform & Neural Net & 137 & 9 & 11 & 4 & 0.588172 & 0.192024 & 0.00494426 \\
normal\_uniform & HGBC & 139 & 7 & 10 & 5 & 0.656313 & 0.346785 & 0.00438808 \\
normal\_uniform & SVM & 141 & 5 & 10 & 5 & 0.663721 & 0.41474 & 0.0033404 \\
normal\_uniform & LR & 134 & 12 & 8 & 7 & 0.675231 & 0.322613 & 0.00376363 \\
normal\_beta & Neural Net & 137 & 9 & 11 & 4 & 0.593275 & 0.202736 & 0.00505157 \\
normal\_beta & HGBC & 140 & 6 & 10 & 5 & 0.658075 & 0.357357 & 0.00437405 \\
normal\_beta & SVM & 142 & 4 & 10 & 5 & 0.666871 & 0.424167 & 0.00297892 \\
normal\_beta & LR & 133 & 13 & 8 & 7 & 0.679679 & 0.328813 & 0.00386298 \\
gamma\_uniform & Neural Net & 137 & 9 & 11 & 4 & 0.593 & 0.202951 & 0.00510945 \\
gamma\_uniform & HGBC & 139 & 7 & 10 & 5 & 0.6597 & 0.357473 & 0.00440051 \\
gamma\_uniform & SVM & 141 & 5 & 10 & 5 & 0.664447 & 0.414572 & 0.00327735 \\
gamma\_uniform & LR & 133 & 13 & 8 & 7 & 0.679753 & 0.328597 & 0.00384738 \\
gamma\_beta & Neural Net & 137 & 9 & 11 & 4 & 0.59558 & 0.209727 & 0.00525216 \\
gamma\_beta & HGBC & 140 & 6 & 10 & 5 & 0.654283 & 0.35311 & 0.00440901 \\
gamma\_beta & SVM & 142 & 4 & 10 & 5 & 0.664786 & 0.417948 & 0.00310447 \\
gamma\_beta & LR & 133 & 13 & 8 & 7 & 0.676524 & 0.32357 & 0.00396213 \\
\bottomrule
\end{tabular}
\end{table}

The simulation study conducted here offered valuable insights into the performance and robustness of various machine learning algorithms across different data distributions. By comparing the original data performance with results from four simulated distributions, we were able to draw significant conclusions about each model's strengths and weaknesses. The simulation study results (see Table \ref{tab:results}) indicate that non-linear models, such as Support Vector Machine (SVM), Histogram-based Gradient Boosting Classifier (HGBC), and Neural Networks, generally outperform traditional linear models  like Logistic Regression in terms of classification metrics such as True Positives (TP), False Positives (FP), True Negatives (TN), False Negatives (FN), area under the receiver operating characteristic curve (AUC-ROC score), and Matthews Correlation Coefficient (MCC). These findings highlight the importance of selecting appropriate machine learning models for agricultural trials, where data can be complex and non-linear relationships are prevalent.

Looking at the original data performance, we find that each model has its unique characteristics. The Neural Net achieved high true positives (136) but also had relatively high false positives (10), resulting in the second-highest AUC-ROC score (0.73242). The HGBC demonstrated excellent precision with the highest true positives (143) and lowest false positives (3), although it had the lowest AUC-ROC score (0.656393) on the original data. The SVM showed a balanced performance with high true positives (142), low false positives (4), and the highest MCC (Matthews Correlation Coefficient) of 0.506071, suggesting good overall performance. Logistic Regression (LR) achieved the highest AUC-ROC score (0.752055) but at the cost of the highest false positives (14), indicating a potential trade-off between sensitivity and specificity.

When we examine the performance across the simulated distributions (normal\_uniform, normal\_beta, gamma\_uniform, and gamma\_beta), several interesting patterns emerge. All models maintained relatively consistent performance across these distributions, which suggests a degree of robustness to different data generation processes. However, we observe that the AUC-ROC scores for the simulated datasets generally decreased compared to those of the original data, with the Neural Net showing the most significant drop (from 0.73242 to around 0.59). This decrease hints at the possibility that the models may be slightly overfitted to the original data distribution.

The MCC scores also saw a decrease for all models in the simulated data, with the Neural Net experiencing the most substantial reduction (from 0.428767 to around 0.2). Interestingly, the SVM consistently maintained the highest MCC across all simulations (around 0.41-0.42), indicating its strong generalization capabilities. The standard error values remained consistently low across all simulations, ranging from about 0.003 to 0.005. These low values indicate high precision in the performance estimates, lending credibility to the conclusions drawn from this study.

Diving deeper into model-specific performance, the Neural Net showed the largest performance drop in simulated data. While it maintained high true positives (around 137), it saw an increase in false negatives (from 7 to 11), suggesting reduced sensitivity. Its consistently lowest MCC in simulations (around 0.2) further indicates potential overfitting to the original data. This performance drop highlights the importance of rigorous testing across various data distributions when deploying neural networks.

The HGBC, on the other hand, demonstrated consistent performance across simulations. It maintained high true positives (139--140) and low false positives (6--7), suggesting good precision and specificity. The stability of its AUC-ROC scores across different distributions (around 0.65--0.66) indicates robustness, making it a reliable choice when the underlying data distribution is uncertain.

The SVM exhibited the most consistent and robust performance across all distributions. It maintained the highest MCC in all scenarios (around 0.41--0.42) and showed a good balance between true positives, false positives, false negatives, and true negatives across simulations. This performance suggests that SVM might be the most reliable choice for this particular problem domain, especially when generalization to unseen data is crucial.

Logistic Regression (LR) showed relatively stable performance across simulations and maintained the highest AUC-ROC scores in simulated data (around 0.67--0.68), suggesting good discrimination ability. However, it consistently had higher false positives (12--13) compared to other models, indicating a potential bias towards positive predictions. This characteristic could be beneficial or detrimental depending on the specific requirements of the application.

The significance of these results extends beyond the performance of individual models. The study underscores the importance of simulation in assessing model generalization and robustness. The differences observed between original and simulated data performance highlight that evaluating models solely on the original dataset may not provide a complete picture of their real-world performance.

Moreover, the relatively consistent performance across different simulated distributions as demonstrated by the low standard errors, suggests that these models are somewhat invariant to the specific probability distribution of the data. This invariance is a desirable property for many real-world applications where the underlying data distribution may be unknown or subject to change, such as data recorded from potato variety trials conducted in a changing climate.

\section{Discussion}
The integration of machine learning algorithms into the potato breeding program has the potential to significantly impact the selection process and overall efficiency. By leveraging the predictive capabilities of these algorithms, breeders can make more informed decisions regarding which clones to advance to future trials, such as the tri-state trials.

In Oregon state-wide trials, the costs associated with producing and distributing seed tubers, field management, and post-harvest grading can exceed \$2,000 per selection. A significant portion of this expense is attributed to grading, as well as data collection and analysis. Data analysis is also time-intensive, as breeders must dedicate considerable time to meetings and discussions aimed at minimizing biases introduced by data from multiple locations. By automating certain aspects of the selection process, machine learning algorithms such as the ones presented here can save time and resources that would otherwise be spent on manual analysis.

Currently, the number of selections tested in two-year Oregon state-wide trials is limited to approximately 50 due to constraints in time and resources. On the other hand, our algorithms enable breeders to assess a larger number of selections in the early stages of trials, effectively addressing bottlenecks. The efficiency gain allows breeders to focus their efforts on more critical tasks, such as evaluating the performance of advanced clones in multi-location trials and making final selection decisions.

The proposed machine learning algorithms, which demonstrated high accuracy in identifying clones to be dropped and moderate performance in predicting clones to be retained, can serve as a valuable tool in the selection process. While these algorithms may not replace the expertise and intuition of experienced breeders, they can complement their decision-making by providing data-driven insights.

By automating certain aspects of the selection process, machine learning algorithms such as the ones presented here can save time and resources that would otherwise be spent on manual analysis. This efficiency gain allows breeders to focus their efforts on more critical tasks, such as evaluating the performance of advanced clones in multi-location trials and making final selection decisions.

Machine learning algorithms can help identify patterns and relationships in the trial data that may not be readily apparent through manual analysis. By considering a wide range of variables and their interactions, the algorithms can provide a more comprehensive assessment of clone performance, taking into account factors such as yield, quality traits, and resistance to environmental stresses. 

The simulation study provides valuable insights into the robustness and generalizability of the machine learning models used in the potato breeding program. Lower performance metrics for the simulated datasets in comparison with the original one indicate slight overfitting to the training data. The variability in neural network performance across different distributions emphasizes the need for careful consideration of data characteristics when applying machine learning techniques. The study also highlights the importance of testing models across various data distributions to assess their performance and reliability. By evaluating the models' performance under different data-generating scenarios, we can better understand their strengths, weaknesses, and potential limitations when applied to real-world scenarios.

However, it is essential to recognize the limitations of machine learning algorithms and computational methods, and emphasize the importance of human expertise in the breeding process. While machine learning can provide valuable predictions and insights, it is ultimately the responsibility of the breeder to interpret the results, consider additional factors, and make final decisions based on their knowledge and experience. We also note that to fully realize the potential of machine learning in the potato breeding program, it is important to continuously refine and validate algorithms using new trial data. Regular updates and improvements to these algorithms can enhance their predictive accuracy and ensure their relevance to the evolving needs of the breeding program.

\section{Conclusion}
We have applied a large collection of machine learning classification techniques to manually collected Russet potato trial datasets. The main goal was to investigate whether and how machine learning methods might provide domain experts with faster and more efficient ways to identify potato varieties for advancement to later trials.

We discovered that although missing data can pose a challenge, we can use imputation or strategic dropping of features, as well as engineer new features, to obtain significant improvements in the model classification performances. As illustrated in the results, the large number of negative examples in our dataset resulted in strong performance for predicting when to drop varieties from the trials. Although this bias in the predictions can lead to false negatives, it also means that varieties that will potentially fail in future trials are dropped earlier on, making the process more efficient and economical. 

Further, our results suggest that integrating machine learning techniques with the art of selection can result in increased selection efficiency. As the variety trials are costly and keeping a clone that is not worthy for one additional year can be cost detrimental. Our results indicate, with availability of extensive data, an integrated art of selection along with data-informed confirmation from machine learning can be effective.

In conclusion, this work has demonstrated the potential of machine learning methods in revolutionizing potato variety selection during trials. The integration of varied data sources, coupled with advanced data-processing techniques, present a pathway toward a more efficient and informed selection process.

\paragraph*{No Acknowledgment: We do not have any funding sources to list.}

\clearpage
\appendix

\section{Preprocessed Dataset Characteristics}\label{tab:potato_feat}
{\bf Appendix Table A1}: Attributes for which data was collected and averaged across 4 replicates of each unique clone.
\begin{longtable}{@{}lp{3.3in}@{}}
    \toprule
        Attribute Name & Attribute Description \\
        \midrule
        Total Yield &  in hundredweight/acre.
Sum of all different sizes, yield 2's and culls where each has been multiplied by 8.31.
Dataset ranges from 0-1510 with mean of 618.
There are 47 missing values.\\
        \hline
         Rank & An integer from 1 to number of varieties for a specific location and year.
It ranks from highest (1) to lowest total yield.
Ranges from 1-44.
There are 47 missing values.\\
         \hline
         \% RB & Total yield divided by Russet Burbank's total yield multiplied by 100.
Ranges from 0-274.
There are 47 missing values.\\
         \hline
         Yield No. 1's & Sum of all yields (in cwt) from which the potatoes are between 4 and 20 ounces.
Ranges from 0-983.
There are 47 missing values.\\
         \hline
         Rank (No. 1's) & An integer from 1 to number of varieties for a specific location and year.
It ranks from highest (1) to lowest yield No 1's.
Ranges from 1-44.
There are 47 missing values.\\
         \hline
         \% RB (No. 1's) & Yield No 1's divided by Russet Burbank's yield No 1's multiplied by 100.
Ranges from 0-336.
There are 47 missing values.\\
         \hline
         \% No. 1's & Yield No 1's divided by total yield multiplied by 100.
Ranges from 0-93.
There are 51 missing values.\\
         $>$ 10 oz. & Yield in cwt of potatoes above 10 oz.
Sums of 10-20 oz (sometimes not provided) and over 20 oz.
Ranges from 0-787.
There are 143 missing values.\\
         \hline
         6-10 oz. & Yield in cwt of potatoes between 6 and 10 oz.
Ranges from 0-716.
There are 144 missing values. \\
         \hline
         4-6 oz.	& Yield in cwt of potatoes between 4 and 6 oz.
Ranges from 0-240.
There are 143 missing values.\\
         Yield No. 2's + $>$ 20oz. & Total yield minus Yield No. 1's minus Yield Under 4 oz.~minus Yield culls.
Ranges from 0-429.
There are 51 missing values.\\
         \hline
         Yield Under 4 oz. & Yield in cwt of potatoes below 4 oz.
Ranges from 0-807.
There are 52 missing values.\\
         \hline
         Yield Culls	& Yield in cwt of discarded potatoes.
Ranges from 0-178.
There are 53 missing values.\\
         \hline
         Yield Over 20 oz. & Yield in cwt of potatoes above 20 oz. There are 140 missing values.\\
         \hline
         Tuber/plant	& Total tuber count divided by the number of plants. There 276 missing values.\\
         \hline
         Average Tuber Size	& Measured in ounces.
Ranges from 0-20.
There are 175 missing values. \\
         \hline
         Length/Width Ratio	& Potato's length divided by its width (computed from raw length and width, which are not separately included among the averages).
Ranges from 0.52-3.5.
There are 19 missing values. \\
         \hline
         Specific Gravity & Potato's air weight divided by the difference of its air and water weights (computed from raw air and water weights, which are not separately included among the averages).
Ranges from 0-1.1.
There are 18 missing values.\\
         \hline
         Fry Color Stem	& Measured in photovolts.
Ranges from 14.4-72.9.
There are 410 missing values.\\
         \hline
         Fry Color Bud & Measured in photovolts.
Ranges from 16.9-76.3.
There are 410 missing values. \\
         \hline
         Hollow Heart & Internal defect variable.
Raw data is an expert-selected number from 1-10, multiplied by 10 to represent a percentage.
Ranges from 0-90.
There are 40 missing values.\\
         \hline
         Brown Center & Internal defect variable.
Raw data is an expert-selected number from 1-10, multiplied by 10 to represent a percentage.
Ranges from 0-30.
There are 40 missing values. \\
         \hline
         Black Spot Bruise	& Internal defect variable.
Raw data is an expert-selected number from 1-10, multiplied by 10 to represent a percentage.
Ranges from 0-87.5.
There are 169 missing values.\\
         \hline
         Internal Brown Spot	& Internal defect variable.
Raw data is an expert-selected number from 1-10, multiplied by 10 to represent a percentage.
Ranges from 0-100.
There are 40 missing values.\\
         \hline
        Vascular Discoloration	& Internal defect variable.
Raw data is an expert-selected number from 1-10, multiplied by 10 to represent a percentage.
Ranges from 0-100.
There are 40 missing values.\\
        \hline
        Flower Color &	Plant characteristic.
Ranges from 1-4 (white to pink).
Variable scale.
Should range from 1-5.
There are 638 missing values.\\
        \hline
        Vine Size &	Plant characteristic.
Ranges from 1.3-5.125.
Variable scale.
Should range from 1-5.
There are 583 missing values.\\
        \hline
        Maturity & Plant characteristic.
Ranges from 0.1375-5.75.
Variable scale.
Should range from 1-5.
There are 639 missing values.\\
        \hline
        Skin Color	& External characteristic.
Ranges from 1.35-71.3.
Variable scale.
Should range from 1-5 (3 is yellow, 4 is brown).
There are 290 missing values. \\
         \hline
        Russeting	& External characteristic.
Ranges from 1-14.625.
Variable scale.
Should range from 1-5.
There are 153 missing values.\\
        \hline
        Tuber Shape	& External characteristic.
Ranges from 1-5 from spherical to long.
There are 261 missing values.\\
        \hline
        Shape Uniformity & External characteristic.
Ranges from 1-5.
There are 153 missing values. \\
        \hline
        Eye Depth & External characteristic.
Ranges from 1-5.5.
Variable scale.
Should range from 1-5.
There are 119 missing values.\\
        \hline
        Greening & External characteristic.
Ranges from 1.625-5.
Variable scale.
Should range from 1-5.
There are 153 missing values.\\
        \hline
        Growth Cracks & External characteristic.
Ranges from 1-5.
There are 153 missing values.\\
        \hline
        Scab & External characteristic.
Ranges from 2.625-5.
Variable scale.
Should range from 1-5.
There are 325 missing values.\\ 
        \hline
        Shatter Bruise	& External characteristic.
Ranges from 2.16-20.83.
Variable scale.
Should range from 1-5.
There are 176 missing values.\\
        \hline
        Bruise Data	& A string of number1 * number2 where number1 is the the diameter of the bruise and number2 is the depth of the bruise.
This attribute is not included in the analysis as there is very little data for it.\\
        \bottomrule
\end{longtable}

\bibliographystyle{plainnat} 
\bibliography{references}

\end{document}